\pdfoutput=1
\documentclass[11pt,hyphens]{article}

\usepackage[]{acl}

\PassOptionsToPackage{hyphens}{url}

\usepackage{times}
\usepackage{latexsym}
\usepackage{epigraph}

\usepackage[T1]{fontenc}
\usepackage[utf8]{inputenc}

\usepackage{microtype}

\usepackage{inconsolata}

\usepackage{graphicx}
\usepackage{paralist}

\title{First Tragedy, then Parse:\\ 
History Repeats Itself in the New Era of Large Language Models}

\author{Naomi Saphra \\
  Kempner Institute at Harvard University \\
  \texttt{nsaphra@fas.harvard.edu}
  \\\And
  Eve Fleisig \\
  University of California - Berkeley \\
  \texttt{efleisig@berkeley.edu} 
  \\\AND 
  Kyunghyun Cho\\
  New York University \& Genentech \\
  \texttt{kyunghyun.cho@nyu.edu} 
  \\\And 
  Adam Lopez \\
  University of Edinburgh \\
  \texttt{alopez@inf.ed.ac.uk}
  }

\begin{document}
\maketitle
\begin{abstract}
Many NLP researchers are experiencing an existential crisis triggered by the astonishing success of ChatGPT and other systems based on large language models (LLMs). After such a disruptive change to our understanding of the field, what is left to do? Taking a historical lens, we look for guidance from the first era of LLMs, which began in 2005 with large $n$-gram models for machine translation (MT). We identify durable lessons from the first era, and more importantly, we identify evergreen problems where NLP researchers can continue to make meaningful contributions in areas where LLMs are ascendant. We argue that disparities in scale are transient and researchers can work to reduce them; that data, rather than hardware, is still a bottleneck for many applications; that meaningful realistic evaluation is still an open problem; and that there is still room for speculative approaches.
\end{abstract}

\section{Introduction}

Picture this scene: A renowned NLP researcher at a hot seven-year-old startup steps onstage to deliver a keynote. The speaker describes an ambitious new system to the packed room, building up to the results slide: a bar chart in which the $x$-axis shows the number of training words, and the $y$-axis shows system accuracy. As each data point is revealed, performance rises relentlessly, culminating in a system trained on well over a trillion words using over a thousand processor cores. It smashes the state of the art by a margin previously thought impossible.

Attendees are visibly shaken as they realize, over the course of a minute, that years of research have just been rendered utterly inconsequential. Established academics panic, anticipating the wholesale rejection of already-submitted grant applications. PhD students despair, contemplating the irrelevance of their unfinished dissertations. Many ponder an exit to industry or a change of fields. They will speak of little else this week.

\begin{figure}
    \centering
    \includegraphics[scale=0.22]{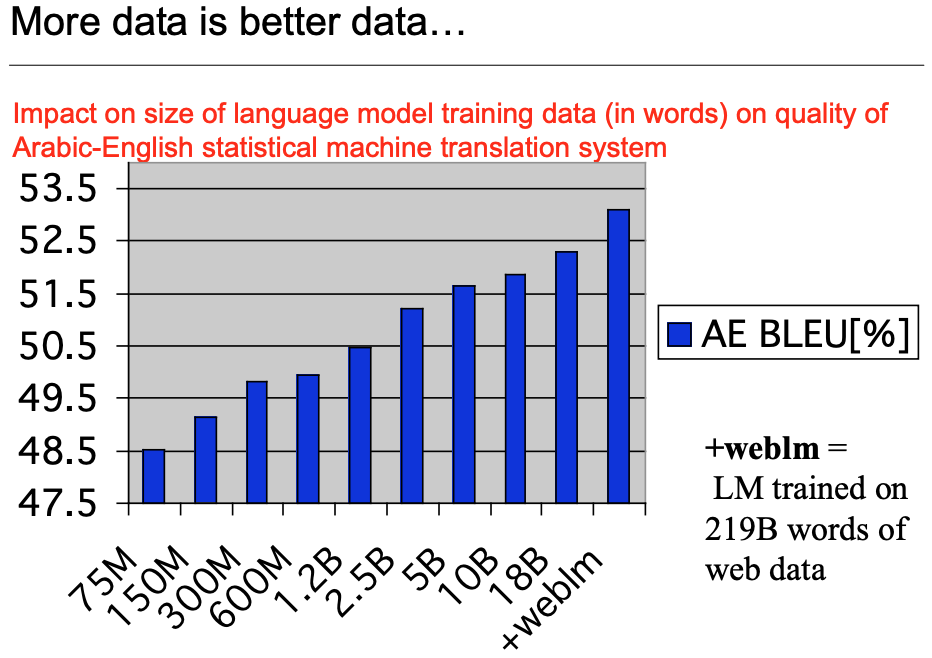}
    \caption{Results slide \citep[reproduced from][]{och-2005-statistical} of Franz Och's keynote talk at the 2005 ACL Workshop on Building and Using Parallel Texts, a predecessor to the Conference on Machine Translation.}
    \label{fig:och_slide}
\end{figure}

Does this scene sound like one that might have happened in the past year? In fact, it happened 19 years ago, in 2005, launching the first era of Large
%\footnote{What qualifies as ``large'' must be relative to the compute and data widely available at any given time. Recall that web-scale data in 2023 inhabits a different order of magnitude from web-scale data in 2005.}
Language Models (LLMs): the \textbf{Statistical Machine Translation (SMT) era}. 
The speaker, Franz Och, had co-invented key methods in SMT \citep{och-ney-2003-systematic,koehn-etal-2003-statistical,och-2003-minimum}, but had not published new work since joining Google in 2004, instead revealing it in an invited talk prior to the launch of a new Google Translate \citep{och_statistical_2006}.\footnote{
The description of the talk and its aftermath is based on the vivid recollections of one of the authors, who was present.}
The provocative results slide from that talk (Figure~\ref{fig:och_slide}) shows how Google improved its SMT system simply by expanding the training corpus of a phrase-based language model \citep{brants-etal-2007-large}.\footnote{
By \textbf{language model} (LM), we mean a generative probabilistic model $\textrm{Pr}(\textbf{x})$ of a string $\textbf{x}$. MT requires a \textit{conditional} LM $\textrm{Pr}(\textbf{y} \mid \textbf{x})$ of target string $\textbf{y}$ given source string $\textbf{x}$. 
In SMT it was originally modeled using the \emph{noisy channel} formulation as $\textrm{Pr}(\textbf{y} \mid \textbf{x}) \propto \textrm{Pr}(\textbf{y}) \times \textrm{Pr}(\textbf{x} \mid \textbf{y})$ \cite{brown-etal-1993-mathematics}. %and later using a discriminative model  \citep{och-ney-2002-discriminative,och-2003-minimum}. 
The \emph{translation model} $\textrm{Pr}(\textbf{x} \mid \textbf{y})$ must be trained on a corpus of example translations, but the LM $\textrm{Pr}(\textbf{y})$ can be trained on \emph{any} data in the target language, making it  amenable to scaling. Like modern LLMs, LMs of the SMT era were generative probabilistic models, albeit based on $n$-grams \citep{shannon1948mathematical} rather than neural networks. Early LMs were widely used across applications, beginning in speech recognition \citep{jelinek-1975-design}, though, unlike contemporary LLMs, they were rarely end products themselves. Although the LM of \citet{brants-etal-2007-large} had a very different architecture from contemporary LLMs, it was an LLM in an important sense: it was trained on 2 trillion tokens, which is comparable to the training data size of modern LLMs.}

The first era of LLMs initially provoked great anxiety among MT researchers about the state of their field, but MT research has continued to flourish in academia, industry, and government. Even in the modern era of deep learning, MT has been a locus of innovations that have fundamentally altered NLP and all of machine learning \citep{bahdanau2015neural,NeurIPS2017_3f5ee243,Sutskever2014SequenceTS}. 

We believe that this history offers  lessons for the \emph{current} era of LLMs, an era during which massive proprietary models have become a de facto baseline for many tasks \cite{anna_rogers_closed_2023}. The expense of state-of-the-art research has led many to question the role of smaller and publicly funded groups in AI \citep{lee2023surveying}, a phenomenon we will call the \textbf{scale crisis}.\footnote{We use the term \emph{crisis} deliberately since others have done so. For example, following the announcement of GPT-4 \citep{openai2023gpt4}, \citet{andriymulyar_andriy_mulyar_my_2023} posted on Twitter that his feed was ``full of ph.d. students having an existential crisis.''} Researchers without direct access to LLMs have publicly fretted over their research directions, with \citet{togelius2023choose} suggesting pivots in research direction to sidestep scale, and \citet{ignat2023phd} sketching research areas that are ``not within the purview of LLMs.'' But what should researchers do if they care about problems that \emph{are} within the purview of LLMs?
To answer this question, we look to the first era of LLMs. What were the durable lessons of that time and evergreen research problems that still matter today? We arrive at several recurrent lessons:

\begin{asparadesc}
    \item \textbf{Scale is supreme}  (Section~\ref{sec:lesson_scale}). We argue that, for areas where data is plentiful, NLP researchers cannot escape the \textbf{Bitter Lesson} \citep{sutton_bitter_2019} that general purpose methods exploiting scale will outperform methods that leverage informed priors. We recommend that researchers take advantage of improvements in hardware as they enable scale at affordable budgets (Section~\ref{sec:hardware}) and that they remember small-scale problems (Section~\ref{sec:small}). 
    
    \item \textbf{Evaluation is a bottleneck} (Section~\ref{sec:lesson_evaluation}). The Bitter Lesson favors generic methods, which require evaluation metrics to optimize over. But improved models create an evaluation bottleneck, since error detection becomes harder when most remaining mistakes are subtle or associated with edge cases. At scale, automated metrics show their flaws. We recommend that researchers work on improving metrics (Section~\ref{sec:metrics}).
    
    \item \textbf{There is no gold standard} (Section~\ref{sec:lesson_human}). When one can afford the annotation costs, it may be tempting to consider human feedback as the ideal solution to the evaluation problem. Unfortunately, history has repeatedly shown that na\"{i}ve methods of soliciting human preferences  result in poor feedback, prioritizing superficial properties of model outputs. This lesson may serve as a counterpoint to the impulse to  collect massive quantities of low-quality data in response to the Bitter Lesson. Instead, we recommend grounding performance measurement in concrete downstream tasks (Section~\ref{sec:concrete}).
    
    \item \textbf{Progress is not continuous} (Section \ref{sec:future}). The ascent of neural MT abruptly ended SMT's decade of seemingly unbeatable growth. This change in directions was enabled by new hardware-based paradigms, so we recommend that researchers continue exploring new methods that might scale well on future hardware (Section \ref{sec:hardware_lottery}).
    
    \item We conclude with a simple message: \textbf{Do research} (Section~\ref{sec:conclusion}). We remind the reader that engineering achievements do not render scientific achievements insignificant, and we encourage the NLP community to renew their commitment to foundational scientific research even in areas where scale is currently a dominant factor.
\end{asparadesc}

\section{Scale is supreme.} 
\label{sec:lesson_scale}

\begin{figure}
    \centering
    \includegraphics[width=0.5\textwidth]{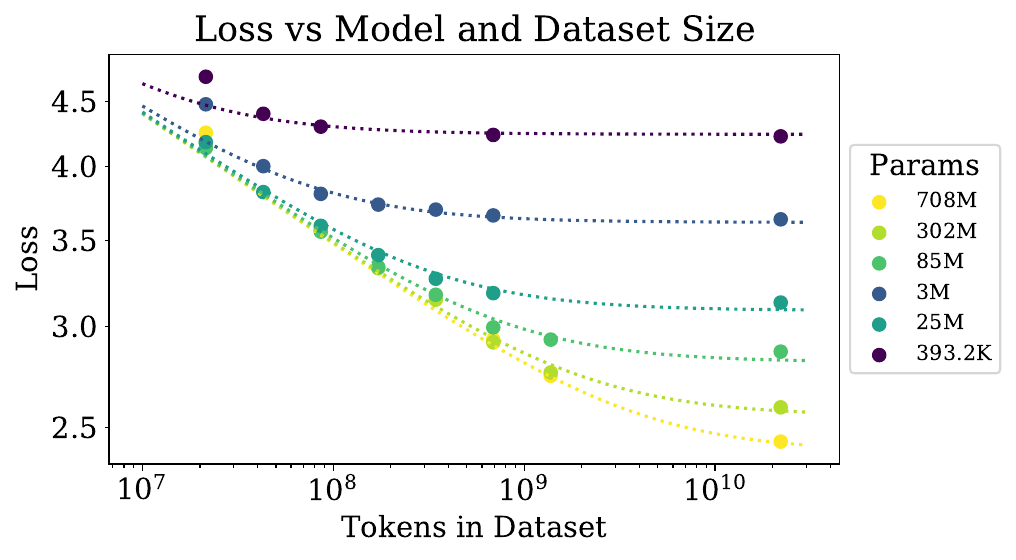}
    \caption{Figure from \citet{kaplan_scaling_2020} illustrating a power law relationship between dataset size and test loss for LLMs with varying numbers of parameters.}
    \label{fig:kaplan}
\end{figure}

The first lesson offered by the history of SMT is that data and compute scale are the dominant factors in system performance. In all eras of MT, improvements in BLEU are logarithmic in training data size \citep{brants-etal-2007-large,koehn-knowles-2017-six}. This is immediately obvious from Figure~\ref{fig:och_slide}: in order to achieve each linear step of improvement in accuracy ($y$ axis), training data size must double ($x$ axis). Figure~\ref{fig:kaplan} \citep[reproduced from][]{kaplan_scaling_2020} shows a strikingly similar log-linear relationship between training data size and system performance for LLMs. Indeed, such relationships are observed across many application areas of machine learning, including vision \citep{mahajan2018exploring} and speech \citep{moore03_eurospeech}.

In a research landscape centered on performance metrics, scale will dominate.
%\footnote{The choice to focus on metrics with log-linear scaling laws may be viewed as a collective decision made by the AI community, rather than a universal requirement. Although many common metrics are theoretically guaranteed to follow scaling laws in some settings \citep{cho_scaling_nodate,hutter_learning_2021}, others empirically differ in their trajectories. Some metrics do not monotonically respond to scale \citep{sagawa_investigation_2020,srivastava_beyond_2022}, or vary significantly based on random factors rather than on scale \citep{sellam_multiberts_2022,mccoy_berts_2019}. These heterogeneous behaviors are elided by aggregate metrics like iid test loss or benchmark averages, which combine diverse subpopulations of data that may exhibit different responses to scale and randomness.  Articulating the limits of scaling laws across evaluation objectives is still an active focus of both theoretical and empirical work.}
 %The resulting access disparities have recently precipitated a crisis in NLP research, where the largest and wealthiest labs dominate conference publications and massive proprietary models have become a de facto baseline for many tasks \cite{anna_rogers_closed_2023}. The expense of state-of-the-art research has led many to question the role of smaller and publicly funded groups in AI. This scale crisis has escalated in response to recent LLM releases. Following the announcement of GPT-4 \citep{openai2023gpt4}, \citet{andriymulyar_andriy_mulyar_my_2023} posted on Twitter that his feed was ``full of ph.d. students having an existential crisis.''
\citet{sutton_bitter_2019} named the resulting malaise the ``Bitter Lesson'': ``General methods that leverage computation are ultimately the most effective, and by a large margin.'' Both SMT and LLMs exemplify this lesson, and in fact Sutton explicitly references natural language processing. Many NLP researchers accordingly feel lost without access to large-scale systems. However, there are limits to scale, and as we will show by reviewing the history of the SMT era, its disparities are often transient. \textbf{The scale crisis is not a permanent state}.

\subsection{Follow the hardware.}
\label{sec:hardware}

%Based on the supremacy of scale, we recommend taking advantage of progress in hardware as it comes, rather than complacently accepting the scale gap as insurmountable.

For several years following the release of Google Translate, large-scale commercial systems dominated the rankings
% \footnote{Shared tasks, which originated in speech recognition and spread to MT, were initially resisted by many researchers. However, they were critical to sustained funding of these fields for the many years during which they were of limited practical use \citep{liberman-2010-obituary} and well established before the first era of LLMs.} 
of translation into English, where data was plentiful \citep{nist_nist_2008,CallisonBurch2009FindingsOT,CallisonBurch2010FindingsOT,CallisonBurch2011FindingsOT,CallisonBurch2012FindingsOT}. Open source tools such as Moses \citep{Koehn2007MosesOS} and collaborations that pooled the resources of small labs narrowed the gap but did not close it until 2013, when translation into English was convincingly won by an academic group using modest hardware \citep{Bojar2013FindingsOT}. The decisive tool was KenLM, an efficient language modeling library \citep{heafield-2011-kenlm,heafield-etal-2013-scalable} that demonstrated how, with the right software, contemporary hardware had made LLM training widely accessible. This end to the SMT scale crisis was the outcome of trends in hardware and software advancement.

The advent of LLMs in the SMT era and their later academic availability were both consequences of \textbf{Moore's law}, a six-decade trend in which computing power has doubled biannually: as \citet{sutton_bitter_2019} observes, ''over a slightly longer time than a typical research project, massively more computation inevitably becomes available.'' SMT-era LLMs  arrived when researchers noticed that they had ignored Moore's law for too long. They rapidly closed the gap: \citet{brants-etal-2007-large} ended a brief race to scale $n$-gram models to web-scale data \citep{Zhang2006DistributedLM,Emami2007LargeScaleDL}. But once the gap was closed, further incremental improvements---which required doubling the training data---necessarily required doubling the hardware cost or waiting for its capacity to double. Well-funded research sought more immediate gains elsewhere, while researchers with longer horizons rode Moore's law towards parity through collaboration and algorithmic advances.
The new era of LLMs has already followed the first part of this pattern: computational requirements of LLMs have been doubling at a rate of less than a year or perhaps faster \citep{sevilla2022compute,dario_amodei_ai_nodate}, much faster than Moore's law. 

The end of the SMT scale crisis was by no means inevitable or foreseeable at the beginning of the first era of LLMs in 2005: it resulted from the efforts of many researchers. We are encouraged to see a similar trajectory forming now. Already, startups advertise cheap large-scale training to the public \citep{portes_mosaicbert_nodate}. Like the groups that competed with commercial translation software in the SMT era, large cross-institutional collaborations are currently pooling resources to build public models \citep{Scao2022BLOOMA1}. A  community has developed around efficient ML, spawning new publication venues like MLSys and developing algorithms already employed in many LLMs \citep{danny_hernandez_ai_2023}.
BERT \citep{devlin-etal-2019-bert}, regarded as inaccessible to many small academic labs at its release, now runs on a consumer-grade M1 MacBook laptop \citep{jared_roesch_octomls_2021}.

In short, small labs do not need to abandon their entire research direction if they are interested in working with state-of-the-art models. Algorithmic efficiency guarantees usually hold across different resource scales, so a method developed on inexpensive hardware can be directly applied at industrial scale. Therefore, \textit{all researchers can seek opportunities to collaborate and develop better algorithms.}
%As long as hardware continues to improve rapidly, data availability will eventually dominate compute availability as the limiting factor on scaling up training. 

\subsection{Remember small-scale problems.}
\label{sec:small}

While directly tackling scale is one strategy, we also recommend pursuing research on problems where data, not compute, is the bottleneck. 

Small-scale settings provide a fertile ground for innovation in data-driven methods; in the previous era of LLMs, SMT researchers often used linguistic structure to improve performance when using smaller data. When such methods showed promise in these development settings, Google Translate inevitably tested them at industrial scale. Nonetheless, at the end of its lifespan, Google Translate’s SMT system remained a phrase-based lookup table. The exploration of classical SMT researchers led to no lasting inventions based on syntax or semantics, and so the assumption that resulting improvements could transfer to large scale settings may have been based on unfounded optimism. Unlike their predecessors, modern NLP researchers preemptively recognize the futility of scaling up data-informed methods, because many of the improvements they offer are already provided by scale. But by leaning too far into the bitter lesson's pessimism now, we risk neglecting settings where, for practical or financial reasons, we \emph{must} learn from limited data. 

When Google Translate launched the first era of LLMs, it was only available between Arabic and English \citep{och_statistical_2006}.
Data was the bottleneck that stood between SMT and its application to many meaningful problems, including, most obviously, the translation between many other language pairs.
Solving this bottleneck required such diverse efforts as the collection of the Europarl corpus \citep{Koehn2005EuroparlAP}, the OPUS corpus \citep{tiedemann-thottingal-2020-opus}, the JHU Bible Corpus \cite{mccarthy-etal-2020-johns} and the Nunavut Hansard \citep{Martin2003AligningAU}; the rapid development of a Haitian Creole corpus in the aftermath of the Haiti earthquake \citep{Lewis2010HaitianCH}; the crowdsourcing of corpora for many Arabic dialects \citep{zbib-etal-2012-machine}; and the development of open-source web crawlers for parallel text \citep{smith-etal-2013-dirt}. It is ongoing today in projects led by groups like Masakhane \citep{adelani-etal-2022-thousand,nekoto-etal-2020-participatory,emezue-dossou-2021-mmtafrica} and 
No Language Left Behind \citep{team2022NoLL}. And yet, MT is still impossible for the vast majority of the world's estimated 7,000 languages.

%For example, \textit{small labs can focus on languages other than English, especially low resource languages like Xhosa or Inuktitut.} NLP researchers can leverage international collaborations with local linguists, incentives outside a profit model, and noncommercial resources to broaden the population that has access to resources in their preferred language. Large-scale systems may continue to dominate the state of the art in English and other widely spoken languages such as Chinese and German, but other researchers have continued to focus on neglected languages in groups like Masakhane NLP \citep{adelani-etal-2022-thousand,nekoto-etal-2020-participatory,emezue-dossou-2021-mmtafrica}. 

Just as no one would have claimed in 2006 that Google had solved all translation problems, no one should claim now that LLMs have solved all NLP problems.\footnote{While we focus on data scarcity in underserved languages, some settings may provide limited data even in English. Such data scarcity may be due to practical hurdles to data collection (e.g., legally protected medical data) or an insufficient profit incentive \citep[e.g. data for speakers of English from lower socioeconomic classes; ][]{curry2024classist} While we cannot say the degree to which pure scale can solve problems like robustness or handling longer contexts, we can identify many problems where data is not collected at scale.} 
An identical bottleneck persists in the current era of LLMs, dominated by anglophone systems like ChatGPT. While these models can handle many languages to some degree due to the incidental multilinguality of any large training corpus \citep{blevins_language_2022}, the training data is overwhelmingly English, and supervision data for learning from human feedback is overwhelmingly from English-speaking Kenyans \citep{billy_perrigo_exclusive_2023}. The hegemony of English has made it a presumed default, inciting the creation of the Bender Rule: ``Always name the language(s) you're working on'' \citep{emily_bender_benderrule_2019}.

Just as MT researchers have done since the SMT era, LM researchers today develop tools which rely less on scale \citep{alabi2022adapting,meyer2022bibletts,park2021klue} for underserved languages. Researchers can leverage international collaborations with local linguists, incentives outside a profit model, and noncommercial resources to broaden the population that has access to technology in their own language. Furthermore, as compute costs continue to decline, even English corpora will become relatively ``low resource'' for future highly overparameterized models. To exploit these datasets more effectively, we turn to a perennial need across AI: quality evaluation metrics.

\section{Evaluation is a bottleneck.}  
\label{sec:lesson_evaluation}

The next lesson offered by reflection on the SMT scale era is that the quality of evaluation methods makes a substantial difference in the effectiveness of training because a good evaluation can be used as a training signal. In SMT, this epiphany was delivered by minimum error rate training, which trained directly on target metrics like BLEU \citep{och-2003-minimum}. Likewise, train-time feedback metrics are often adapted for test-time evaluation: language modeling work may present validation loss, or equivalently perplexity, as the direct measurement of language modeling performance. It is therefore easy to use \textbf{symmetric} evaluations, %\footnote{Note that any evaluation method applied symmetrically to both train and test threatens to become a particularly egregious exemplar for \textbf{Goodhart's law} \citep{goodhart1984problems}: ``When a measure becomes a target, it ceases to be a good measure'' \citep{strathern1997improving}. This caveat questions the validity of any proxy for quality we might use to compare systems, but the law is particularly salient when we train on our test metric.} 
applying the same metric for training feedback and test-time performance assessment.
Like model evaluation, training can be based on comparison with a ground truth, as in conventional training; quality estimation based on output alone, as used often in Reinforcement Learning (RL) settings \citep{konda_actor-critic_1999,pmlr-v32-silver14,bai2022constitutional}; or direct feedback, as provided by RL from Human Feedback (RLHF) and related methods of human assessment \citep{ ziegler2019finetuning, stiennon2020learning, ouyang2022training, bai2022training}. 
The evaluation metric then becomes a crucial lever to improve model quality.

With large scientific projects in both industry \citep{noauthor_stability_nodate,thoppilan_lamda_2022} and nonprofit sectors \citep{biderman2023pythia,Scao2022BLOOMA1} spending millions on training LLMs, it may be surprising to point away from scale to other bottlenecks in system performance. The focus on scale bottlenecks is reasonable, as for any given compute budget, there is an optimal quantity of training data that yields the most accurate model \citep{Hoffmann2022AnEA}. Either compute or data size, therefore, can become a limiting factor, and there are still financial barriers to compute \citep{lee2023surveying} and limits on how much unstructured natural language data is available \citep{villalobos2022will}. However, training requires data and compute resources to be connected by evaluation feedback; this connection determines the effectiveness of training. 
% 
% In its most general formulation, evaluation can be thought of as a reward function $r$, operating on a model output $y$ and some global context $C$ representing the resources available for evaluation:
% \begin{equation}
%     r(y, C)
% \end{equation}
% Evaluation can be automated or human in its source; unidimensional or complex in form; and obtained through direct feedback or comparison to some ground truth. 

% Training itself is also an act of repeated evaluation. For example, cross-entropy loss can be formulated as evaluation with a reward function using a ground truth label $y^*$ as global context:
% \begin{equation}
%     r(y, \{y^*\}) = \sum_{i} y_i \log y^*_i
% \end{equation}
% When training instead with direct human feedback such as a Likert score, the reward function's context is the human feedback mechanism and the reward is directly expressed as the score assigned by human feedback. Therefore, 
% Evaluation during training can be based on comparison with a ground truth, as in conventional training; quality estimation based on the output alone, as used often in RL settings \citep{konda_actor-critic_1999,pmlr-v32-silver14,bai2022constitutional}; or direct feedback, as provided by RLHF and related methods of human assessment \citep{ ziegler2019finetuning, stiennon2020learning, ouyang2022training, bai2022training}. 

Unfortunately for efficient automatic approaches to evaluation, automatic metrics often fail to predict human evaluation, as shown for language modeling loss \cite{liu_same_2023} and BLEU \cite{reiter_structured_2018}. Even a metric that mirrors the training objective slightly can artificially inflate model performance; for example, it was observed early on that a phrase-based evaluation metric like BLEU might favor phrase-based SMT \citep{riezler2005some}. Identifying good metrics is challenging and becomes more difficult with each improvement, as the remaining errors become increasingly subtle or complex. This challenge was also recognized early in SMT, with calls for a ``BLEU++'' \cite{och-2005-statistical}.

Evaluation, therefore, has become a crucial goal for modern LLM research. While training often relies on cross-entropy loss or other simple comparisons between each token in a sequence, benchmarking a trained LLM typically uses different evaluation criteria, such as checking the final answer in a word problem or the accuracy of a prompt-based classifier \citep{laskar2023systematic}. However, these tests are plagued by data contamination: benchmark exposure during training has created illusory gains in tasks ranging from code generation \citep{khan_xcodeeval_2023} to theory-of-mind puzzles \citep{ullman2023large}. Clearly language models are improving, but we cannot say precisely how, or by how much.

\subsection{Improve the metrics.}
\label{sec:metrics}

Because evaluation is a bottleneck, we recommend greater focus on improving metrics, a goal which can yield rewards even with limited access to scale. Straightforward increases in computational infrastructure and raw data collection yield predictably diminishing returns on investment. Evaluation, by contrast, provides a conceptual space that can reward innovation and careful work with new insights and unknown improvements in system capabilities. What are the fundamental problems in this space, and how might we approach them? 

One argument for why automatic metrics and static benchmarks are poor methods of evaluation is that they fail at measurement modeling; that is, these metrics do not actually measure what they purport to measure. This concern is reflected in objections to benchmarks and metrics that fail to reflect human evaluation \cite{liu-etal-2016-evaluate, novikova-etal-2017-need} or improvement on natural language understanding more broadly \cite{raji2021ai}. 
These discussions parallel the unease in the MT community when studies found that automated metrics such as BLEU did not always correlate with human judgments \cite{callison-burch-etal-2006-evaluating}. In response, SMT saw a flurry of incrementally improved bitext-based metrics \citep{stanojevic_fitting_2014,popovic_chrf_2015,mutton_gleu_2007}. In a scale crisis, improving evaluation metrics that leverage naturally available data like bitext can be a worthwhile focus.

Unfortunately, the naturally available data used in evaluation can still contaminate training corpora. Furthermore, automated metrics that rely on a static ground truth cannot reflect general quality \citep{raji2021ai}, model conditions under interactive deployment, or provide on-policy reward feedback for reinforcement learning. These issues motivate automated metrics that do not require ground truth, although proposals based on AI supervision are themselves difficult to evaluate due to the same issues of dataset bias and contamination.

% The field of quality estimation in MT focuses on evaluations that don't rely on ground truth bitext. \ns{cite} Similarly, evaluation based only on properties of the output is the holy grail of Reinforcement Learning With AI Feedback (RLAIF), making improvements in unsupervised evaluation metrics an urgent goal for modern LLM applications. \ns{what's the point here}

Despite a research community strongly motivated to improve them, even the best automated metrics are far from perfect. As automated metrics and static benchmarks fail, researchers with resources are increasingly hiring humans to assess model outputs. Facing similar challenges in SMT, researchers also called for human evaluation to be prioritized, both for benchmarking \cite{callison-burch-etal-2006-evaluating,CallisonBurch2007MetaEO} and for training \citep{hopkins-may-2011-tuning}. However, human evaluation does not intrinsically solve problems with measurement modeling, and raises challenges of its own.

\section{There is no gold standard.}  \label{sec:lesson_human}

Language evolved to be interpreted by humans. This fact leads us to a tempting myth: that we can easily evaluate synthetic natural language outputs by simply asking a human for their opinion. To the contrary,  the next lesson we discuss from the SMT era is that human annotation cannot provide a universal ``gold standard'' of quality feedback. When MT competitions proudly turned to human evaluation as the highest-quality and most reliable option for choosing a winner, critics pointed out that these evaluations failed the basic expectations of consistency needed for a fair ranking \citep{bojar2011grain,Lopez2012PuttingHA}. Even soliciting useful, let alone perfect, evaluations from humans turned out to require careful thought and trade-off decisions.

Practitioners often rely on naive methods of soliciting human feedback on LLM outputs, such as single rating scales or ranking model outputs by quality, which do not distinguish why annotators prefer a particular model output and thus offers limited guidance. For example, OpenAI’s ChatGPT annotator interface asks the user to rank outputs from ``best'' to ``worst'' \cite{ouyang2022training}.  Anthropic adds extra dimensions but with limited guidance, as annotators evaluate the extent to which generated text is ``helpful'' and ``harmful,'' claiming that the vagueness of these guidelines permits versatile human preferences \citep{bai2022training}.  Recent work goes even further by using freeform text feedback, rather than predefined numerical axes \citep{shuster_blenderbot_2022,lnls-2022-learning,scheurer2022training,scheurer2023training}, although how best to incorporate these explanations remains an unsolved problem.

The NLP community, however, is rediscovering that eliciting human preferences without clear guidance produces data that is not only noisy, but introduces systematic errors in models trained on the data. When many dimensions of quality are collapsed into a single preference scale, outputs that are worse along some dimensions may have higher ratings because they perform well along others. In these cases, annotators prioritize fluency over other aspects of the text, such as factuality or consistency \cite{clark-etal-2021-thats}. LLMs consequently prioritize fluency of large language model outputs over factuality \cite{ji2023}, mirroring concerns from the SMT era that models prioritized fluency over the faithfulness of translations \cite{Dorr2011}---concerns that have since been empirically confirmed \citep{Martindale2018FluencyOA}. \citet{belz_hastie_2014} and \citet{vanderlee2021} note that overall quality of generated text is often ``too abstract'' to be measured and both recommend the use of separate criteria for different dimensions of the text to distinguish what specific issues are present in a model output. \citet{gerhmann2023} and \citet{vanderlee2021} warn that vague annotation guidelines can exacerbate annotator confusion, underscoring the importance of clearly defining the different dimensions on which to rate text quality.

Even after these refinements to the evaluation process, human evaluation for MT has encountered issues that remain unsolved, and current research suggests that evaluation of current models will increasingly encounter similar issues, including the following challenges.

\paragraph{Specifying evaluation criteria is hard.} 
Even when evaluation criteria are separated into several axes, these scores are correlated, suggesting that human evaluators have difficulty in separating out criteria such as adequacy and fluency \cite{novikova-etal-2018-rankme}. In addition, many studies fail to define their axes \cite{vanderlee2021}, permitting evaluators to differ even more in their interpretation of the task and thus increasing variation among annotators. That is, separating out axes of evaluation is necessary but not sufficient to identify multiple desirable traits of model output, a recurring problem in the history of MT evaluation \citep{Chatzikoumi2019HowTE}. Even when evaluation criteria can be defined clearly, crowdsourced annotators often lack the necessary expertise to follow them. Crowdworkers therefore align poorly with expert annotators, even underperforming against automatic evaluation metrics \citep{freitag_experts_2021}. Some problems with objective specification can be resolved by defining multiple objectives and consulting expert annotators rather than crowdworkers.

\paragraph{Individual preferences are inconsistent.} 
Classic SMT results reveal another fundamental problem in human evaluation: pairwise human rankings often fail to produce a consistent order \citep{bojar2011grain,Lopez2012PuttingHA}. Any approach based on comparing outputs therefore reflects an unrealistic expectation of consistency in human preferences. The signal provided by ranking is noisy.

\paragraph{Disagreement isn't just noise.} 
When human evaluators disagree on the quality of text, this does not necessarily reflect ``noise'' or ``random variation'' but rather genuine differences in opinion among evaluators \citep{larimore-etal-2021-reconsidering, Patton2019AnnotatingSM, prabhakaran-etal-2021-releasing, pavlick2019,basile-etal-2021-need,plank-2022-problem}, a problem that has long plagued MT evaluation \cite{lommel2014assessing}. The management of diverse annotator preferences is only exacerbated when benchmarking and training on freeform text from varied sources \citep{giulianelli-etal-2023-comes}. Furthermore, aggregation of annotator judgments obscures the opinions of underrepresented groups \cite{prabhakaran-etal-2021-releasing,fleisig-etal-2023-majority}, and use of inter-annotator agreement as a quality metric causes additional erasure of perspectives by denying that these priorities are contested \cite{blodgett2021thesis}.

\paragraph{} 
These issues collectively prevent human evaluation from providing clear feedback on model outputs. The fact that they have remained major concerns in MT despite decades of research suggests that current researchers would do well not to underestimate the challenges posed by these issues. Furthermore, issues of both task specification and disagreement may be even more central to the evaluation of current models that can handle more varied tasks. Whereas fluency and faithfulness to a source text might cover major concerns in MT, there is a broader range of criteria that generated text must fulfill, such as informativeness and coherence \cite{vanderlee2021}. These requirements, along with increased freedom to produce text on topics where there is real-world disagreement, including social, ethical, and political concerns \cite{abid2021, blodgett-etal-2020-language, Liu_Jia_Wei_Xu_Wang_Vosoughi_2021, zhao2021ethicaladvice}, mean that human evaluation issues will be pressing problems for the NLP community to solve.

\subsection{Focus on concrete tasks.}
\label{sec:concrete}

Due to the inherent flaws of evaluation based on human assessment, we recommend measuring concrete tasks under deployment conditions. Extrinsic evaluations \cite{belzreiter2006}, wherein model output quality is evaluated based on utility for specific downstream applications, are still uncommon in evaluation of text generation \cite{vanderlee2021}. However, they may be more useful for evaluating the quality of content or meaning \cite{reiterbelz2009,ehudreiter_future_2023} because human assessment often fails to predict performance on downstream applications \citep{kunz-etal-2022-human}.

In MT, concrete 
downstream objectives have long been used in evaluation. \citet{snover-etal-2006-study} examined how many manual edits human translators had to make to model output, reflecting the desiderata of human-AI collaboration settings. Other metrics rely on the user's ability to accomplish specific tasks using model output, such as answering reading comprehension questions based on translations \citep{jones-et-al-2005-measuring,callison-burch-2009-fast,scarton-specia-2016-reading} or summaries \citep{wang2020asking}. A recent and growing body of research attempts to measure the effectiveness of MT in second language education \cite{lee2023effectiveness}.

In general, the best evaluations are likely to rely on realistic assessment of what LLMs enable humans to do. In modern LLMs, work on the challenges of evaluation is likely to draw on insights from human-computer interaction.\footnote{Interdisciplinarity between machine learning applications and HCI is a perennial concern. HCI researchers are periodically invited to speak at ML and ML applications conferences, e.g., NeurIPS hosted HCI-centered keynotes from Deborah Estrin in 2013 and Juho Kim in 2023.} Good user trials require careful study design and consideration of human variety, as well as an understanding of individual psychology.

\section{Progress is not continuous.}
\label{sec:future}

Our final lesson is that new paradigms can unlock new orders of scale and even new scaling coefficients, leading to abrupt improvements in performance. The SMT era, fueled by large $n$-gram models, lasted for over a decade, with scale providing increasing improvements over time. But Moore's law was threatened during this era due to the breakdown of Dennard scaling, the observation that smaller transistors require commensurately less power, meaning that they can be miniaturized while keeping power consumption constant.
To drive continued improvement, hardware manufacturers turned to parallelization. Graphical processing units (GPUs), which favor high parallelization of code with minimal branching---and thus simpler and smaller processors---were soon being repurposed to train neural networks \citep{Hooker2020TheHL}.

Neural networks had been investigated in SMT for years. Indeed their earliest use in SMT was as $n$-gram language models, when \citet{schwenk2006continuous} built an SMT decoder using the neural $n$-gram model of \citet{Bengio2003ANP}---an idea that only began to gain traction almost a decade later when revisited by  \citet{devlin-etal-2014-fast}, whose ACL best paper award marked a shift in NLP establishment attitudes towards neural networks. The increasing power of GPUs and their use in training neural networks fueled new research in end-to-end neural MT (NMT), enabling \citet{Kalchbrenner2013RecurrentCT} to revive the even older idea of an encoder-decoder architecture \citep{eco1997AsynchronousTW}. Advances such as  attention \citep{bahdanau2015neural}, seq2seq \citep{Sutskever2014SequenceTS}, and transformers \cite{NeurIPS2017_3f5ee243} followed in a flurry of activity. Within two years, NMT swept the annual shared tasks \citep{jean-etal-2015-montreal,chung-etal-2016-nyu,bojar-etal-2016-findings}, and in 2016, Google Translate announced that it had switched to NMT \citep{Wu2016GooglesNM,barak_turovsky_found_2016}. 
Research on SMT quickly faded.

GPUs effectively introduced a new dominant \textbf{paradigm} by creating conditions that favored deep learning. \citet{Kuhn1962} described scientific advancement as a cycle of scientific revolutions in which paradigms such as phrase-based SMT or deep learning emerge, followed by periods of \textbf{normal science} when researchers aim to apply, articulate, and expand the fact base of the paradigm. Often, scientific revolutions result from the availability of new tools. \citet{Hooker2020TheHL} drew on this framework of scientific revolution to analyze the landscape of AI research, identifying the \textbf{Hardware Lottery} as a situation in which hardware dictates methods.\footnote{\citet{gururaja2023build}, whose oral history of NLP---including comments on the current scale crisis and the cyclic nature of what they call exploit-explore incentives---complements our work, also point to similar paradigm shifts in NLP emerging from a \textit{software} lottery.} 
Under the Hardware Lottery, GPUs offered a winning ticket for deep learning to reshape MT. What research objectives are recommended by the resulting NMT revolution? 

\subsection{Shape the hardware.}

\label{sec:hardware_lottery}

The Hardware Lottery tells us that hardware guides the direction of research, but researchers can also direct the design of hardware. While  these new tools may enable scientific revolutions, \citet{Kuhn1962} pointed out that the development of new tools is itself shaped by the reigning paradigm and by the normal scientific process. Hardware design itself is an example, having been driven for many years by incremental improvements to a paradigm of miniaturization and parallelization of transistors. But hardware manufacturers are approaching the physical limits of miniaturization, and the path forward is again uncertain \cite{Lundstrom2022MooresLT}, as it was at the end of Dennard scaling . Therefore, we recommend that researchers focus not only on developing and using new hardware, but on anticipating potential hardware developments and developing algorithms for platforms before they are widely available.

By creating software tools and algorithms that can take advantage of hardware designed for sparsity \citep{krashinsky_nvidia_2020} or new sources of parallelism \citep{launay_hardware_2020}, researchers can develop techniques preemptively for future technologies. At the same time, they also create a market to motivate the development of new hardware that can enable the next revolutionary development. Researchers may even co-design hardware and software jointly, a strategy likely to drive future computing advances \citep{Leiserson2020TheresPO,Lundstrom2022MooresLT}. It is the possibility of reshaping tools for the future that makes alternative paradigms worth exploring under a scale crisis.

\section{Conclusion: Do research.}
\label{sec:conclusion}

As pure engineering efforts and institutional wealth outstrip novel scientific work, some in the AI community are pessimistic about the prospects of foundational research. Our position, articulated over the course of this paper, is that there is much exciting, timely work yet to be done. 

These lessons are not particular to LLMs, but apply to any field subject to the Bitter Lesson. For example, issues in human evaluation plague many disciplines in machine learning. In computer vision, annotator idiosyncrasies account for many of the remaining inaccuracies of modern ImageNet models~\citep{shankar_evaluating_2020}. If a constrained labeling task such as image classification is subject to varied human judgment \cite{parrish_is_2023}, how much harder is it to annotate free text generation?

Beyond our specific recommendations for researchers interested in improving the capabilities of language models, we would also point to scientific opportunities across related fields. From interpretability to empirical training analysis to public policy, many research areas only become more relevant and complex as models rapidly improve. Furthermore, while novel modeling work suffers in a scale crisis, we can focus on new architectures and algorithms that take advantage of existing hardware and even anticipate future tools.

Some speculative alternatives to phrase-based SMT, such as explicitly modeling syntax \citep{galley2004s,Collins2005ClauseRF,Chiang2007HierarchicalPT} and semantics \citep{Jones2012SemanticsBasedMT}, were obviated by the Bitter Lesson and its expression in NMT. However, other proposals  formed the basis of the NMT era. Many enduring  careers in NLP research were forged in areas that are now forgotten, and without risky exploration of unproven directions, the field could not have achieved many breakthroughs. Our anxieties should not discourage us from seizing the opportunities presented by a new era of LLMs.

\section*{Limitations}

The positions taken in this paper are based on both experience and reading of historical trends in natural language processing. While we believe that the lessons we identify in this paper are durable, history does not always repeat, and our oracular powers are otherwise limited. Even after considering our position, researchers should use their own best judgement on directions to pursue.

\section*{Ethical Considerations}

The authors received permission from Andriy Mulyar to feature his Twitter post as an example of the March 2023 scale crisis discourse.

\section*{Acknowledgements}
We thank Kenneth Heafield, Nikolay Bogoychev, and Steven Kolawole for helpful discussion; and Arya McCarthy, Kevin Yang, Sanjay Subramanian, and Nicholas Tomlin for comments on previous drafts. We thank our reviewers for their constructive feedback.

This work was supported by Hyundai Motor Company (under the project Uncertainty in Neural Sequence Modeling) and the Samsung Advanced Institute of Technology (under the project Next Generation Deep Learning: From Pattern Recognition to AI). This work has been made possible in part by a gift from the Chan Zuckerberg Initiative Foundation to establish the Kempner Institute for the Study of Natural and Artificial Intelligence. During the majority of work on this paper, Naomi Saphra was employed by New York University.

\bibliography{custom}

\end{document}